\documentclass[journal]{IEEEtran}

\usepackage{array}
\usepackage{amsmath,amsfonts}
\usepackage{bm}
\usepackage{booktabs}
\usepackage{cite}
\usepackage{fancyhdr}
\usepackage{graphicx}
\usepackage{hyperref}
\usepackage{makecell}
\usepackage{stfloats}
\usepackage[caption=false,font=normalsize,labelfont=sf,textfont=sf]{subfig}
\usepackage{textcomp}
\usepackage{threeparttable}
\usepackage{url}
\usepackage{verbatim}

\hypersetup{colorlinks=true, linkcolor=blue, anchorcolor=blue, citecolor=blue, urlcolor=blue, filecolor=bluu}
\begin{document}

\title{BA-Net: Bridge Attention in Deep Neural Networks}

\author{Ronghui Zhang, \and Runzong Zou, \and Yue Zhao, \and Zirui Zhang, \and 
        Junzhou Chen, \and Yue Cao, \and Chuan Hu, \and Houbing Song}

\IEEEpubid{0000--0000/00\$00.00~\copyright~2021 IEEE}

\maketitle

\begin{abstract}
Attention mechanisms, particularly channel attention, have become highly influential in numerous computer vision tasks. Despite their effectiveness, many existing methods primarily focus on optimizing performance through complex attention modules applied at individual convolutional layers, often overlooking the synergistic interactions that can occur across multiple layers. In response to this gap, we introduce bridge attention, a novel approach designed to facilitate more effective integration and information flow between different convolutional layers. Our work extends the original bridge attention model (BAv1) by introducing an adaptive selection operator, which reduces information redundancy and optimizes the overall information exchange. This enhancement results in the development of BAv2, which achieves substantial performance improvements in the ImageNet classification task, obtaining Top-1 accuracies of 80.49\% and 81.75\% when using ResNet50 and ResNet101 as backbone networks, respectively. These results surpass the retrained baselines by 1.61\% and 0.77\%, respectively. Furthermore, BAv2 outperforms other existing channel attention techniques, such as the classical SENet101, exceeding its retrained performance by 0.52\% Additionally, integrating BAv2 into advanced convolutional networks and vision transformers has led to significant gains in performance across a wide range of computer vision tasks, underscoring its broad applicability. 
\end{abstract}


\section{Introduction}
\IEEEPARstart{T}{hroughout} the past decade, Convolutional Neural Networks (ConvNets) have been extensively utilized as a dominant approach for a wide range of computer vision tasks. Since the introduction of AlexNet \cite{krizhevsky2012imagenet}, a significant body of research has concentrated on refining ConvNet architectures to boost their overall performance. These continuous advancements have laid the foundation for more sophisticated models that excel in both accuracy and efficiency across various domains \cite{simonyan2014very, he2016deep, huang2017densely}.

In contrast to efforts aimed at enhancing the overall structure of ConvNets, a substantial body of research has shifted its focus towards the introduction of attention mechanisms. Channel attention, a key component of attention mechanisms, assigns attention weights to features derived from neighboring convolutional layers to amplify relevant information while suppressing less important features. A well-known example of channel attention is the SE (Squeeze-and-Excitation) module \cite{hu2018squeeze}, which utilizes global average pooling (GAP) to compress the information from each channel and then employs fully-connected layers followed by a Sigmoid activation function to generate channel weights in the excitation phase. However, relying solely on GAP for encoding channel information limits its ability to fully capture complex feature relationships. To address this limitation, several methods have been proposed to enhance the SE module by generating more intricate channel-wise representations. For instance, CBAM \cite{woo2018cbam} and SRM \cite{lee2019srm} improve upon GAP by incorporating global max pooling and global standard deviation pooling, respectively. Furthermore, approaches like GSoPNet \cite{gao2019global} and FcaNet \cite{qin2021fcanet} replace GAP with more advanced techniques such as global second-order pooling and discrete cosine transform (DCT). Additional research has explored the integration of complementary attention mechanisms, such as spatial attention \cite{woo2018cbam}, coordinate attention \cite{hou2021coordinate}, and self-attention \cite{wang2018non, cao2019gcnet, bello2019attention}, with the SE module to achieve further improvements in performance.

Much of the existing research focuses on extracting enhanced attention from individual convolutional layers. However, this approach frequently overlooks the potential advantages of integrating outputs across multiple layers, which can capture richer feature representations. To compensate for this limitation, many methods employ increasingly complex strategies or introduce additional network branches to boost performance. While these approaches may improve accuracy, they also significantly increase the model's overall complexity, making it more computationally demanding. Striking a balance between performance gains and model efficiency remains a key challenge in attention mechanism design.

\IEEEpubidadjcol

To address the aforementioned challenges, we propose a novel approach that bridges different convolutional layers to generate more effective attention. When transmitting information between convolutional layers, issues such as signal loss and feature degradation are inevitable. By bridging features from earlier convolutional layers, valuable attention information can be preserved and enhanced. Furthermore, due to the stacked structure of convolutional layers, there exists an implicit correlation between the attention mechanisms of the current and preceding layers. Prior works, such as ResNet \cite{he2016deep} and DenseNet \cite{huang2017densely}, have already highlighted the critical role of linking previous features in deep neural networks. DIANet \cite{huang2020dianet} extends this concept by introducing a framework that shares an attention module across multiple layers, promoting the integration of layer-wise information. Similarly, the recently developed BViT \cite{li2023bvit} further demonstrates that incorporating attention relationships across layers can significantly boost the performance of vision transformers.

Building on the preceding discussion, we enhance the SE module by incorporating features from prior layers into the attention mechanism, resulting in the development of a novel Bridge Attention (BAv1) module \cite{zhao2022ba}. The BAv1 module compresses features from both previous and adjacent convolutional layers using global average pooling (GAP) and integrates them through a bridging operation to produce more informative channel representations. This approach leads to the generation of improved channel attention weights. In this paper, we introduce adaptive selection operators within the BAv1 framework, allowing for the creation of selection weights for features from various layers, which we denote as BAv2. As a result, our proposed BAv2 module is capable of selectively learning from multiple layers during the aggregation process, thereby enhancing both performance and robustness, all while maintaining a lightweight and flexible design that facilitates seamless integration into ConvNets.

Furthermore, our research goes beyond merely applying the BAv2 module to convolutional networks; it also encompasses vision transformers. Since the introduction of Vision Transformer (ViT) \cite{dosovitskiy2020image}, a growing body of literature has emerged that investigates various variants and achieves significant advancements \cite{touvron2021training, wang2021pyramid, wu2021cvt, liu2021swin, wang2022pvt}. Much of the current research is concentrated on designing more sophisticated transformer backbones, a task that presents considerable engineering challenges. In contrast, our approach integrates the proposed BAv2 module with state-of-the-art vision transformers, thereby effectively enhancing their overall performance. This integration not only improves accuracy but also maintains the efficiency and adaptability of the existing transformer architectures.

In summary, our major contributions can be summarized as follows.

\begin{enumerate}
    \item \textbf{Innovative Bridge Attention Module:} To address the limitations of traditional channel attention mechanisms, we introduce the innovative Bridge Attention (BA) module. This approach enhances information utility by integrating features from earlier layers, thereby improving attention accuracy. Additionally, BAv2 has been designed to adaptively optimize feature weights across layers, further refining the attention generation process.
    
    \item \textbf{Enhancing Performance Across Diverse Architectures:} In response to the demand for broader applicability of attention mechanisms, the BAv2 module has been effectively integrated into various advanced deep neural network architectures. This integration enhances performance in both convolutional networks and vision transformers, setting a new benchmark for the implementation of channel attention in complex neural networks.
        
    \item \textbf{Improving Training and Evaluation Standards:} To address the challenges posed by inconsistent training standards in evaluating attention mechanisms, our team has refined and standardized training methodologies. This approach facilitates robust and fair comparisons with existing methods, establishing a contemporary benchmark for performance assessments. Extensive validation across various computer vision tasks confirms that the BAv2 module consistently outperforms established architectures.
\end{enumerate}

\begin{figure}[!t]
\centering
\includegraphics[width=0.45\textwidth]{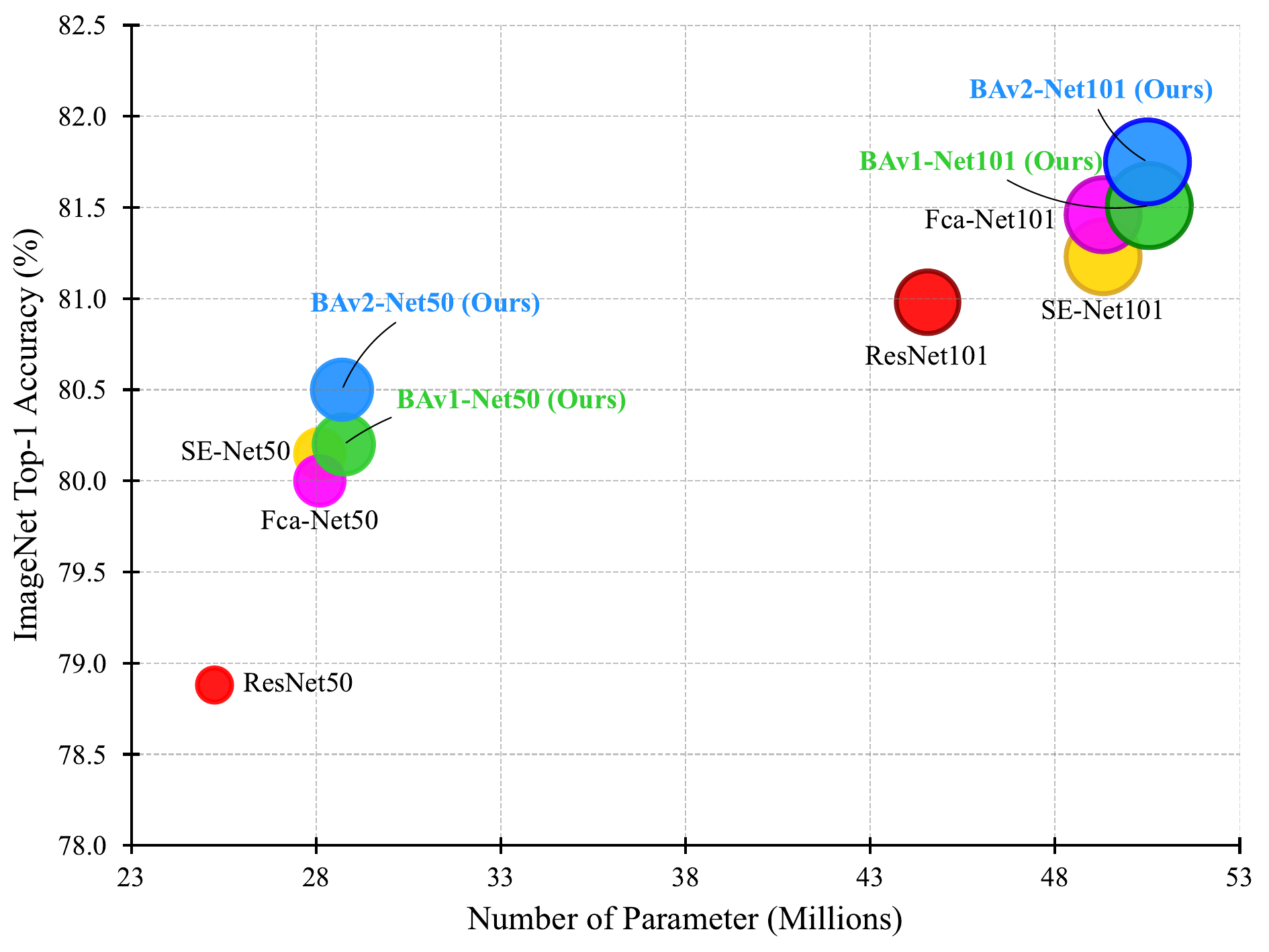}
\caption{Comparison of state-of-the-art attention modules (SENet \cite{hu2018squeeze}, FcaNet \cite{qin2021fcanet}, BAv1 \cite{zhao2022ba} and BAv2) applied on ImageNet \cite{russakovsky2015imagenet} dataset with ResNets \cite{he2016deep} as backbones. The evaluation criteria include accuracy, network parameters and FLOPs. The size of circle represnets the FLOPs.}
\label{fig_1}
\end{figure}

We conducted extensive experiments on several widely recognized datasets, consistently achieving state-of-the-art performance. As shown in Fig. \ref{fig_1}, our BAv2 module, integrated with ResNet \cite{he2016deep}, outperforms existing attention methods on ImageNet \cite{russakovsky2015imagenet}. Additionally, in both CIFAR-10 and CIFAR-100 \cite{krizhevsky2009learning}, our approach exceeds the performance of competing attention techniques across various backbone architectures, demonstrating the broad applicability of the BAv2 module. We also applied the BA module to advanced deep neural network architectures, including ResNeXt \cite{xie2017aggregated}, RegNet-Y \cite{radosavovic2020designing}, PVT v1 \cite{wang2021pyramid}, Swin Transformer \cite{liu2021swin}, CSWin-Transformer \cite{dong2022cswin}, YOLOv8 \cite{Jocher_Ultralytics_YOLO_2023}, and YOLOv9 \cite{wang2024yolov9}. The results validate the effectiveness of our proposed module in both convolutional networks and transformers. Experimental evaluations on the COCO \cite{lin2014microsoft} dataset further confirm that the BAv2 module is highly effective and adaptable for enhancing performance across various downstream tasks.

This manuscript represents a significant expansion of our previous work, as outlined in \cite{zhao2022ba, zhao2021ba}. Key advancements include the sophisticated integration of an adaptive selection operator within the BAv1 module, along with the adaptation of channel attention mechanisms from traditional convolutional networks to advanced vision transformers. In the experimental section, we meticulously retrained our methodology alongside established attention techniques using state-of-the-art training methods, ensuring that our evaluations are both current and rigorously fair. Extensive experiments and detailed analyses have been conducted, particularly focusing on small-scale datasets and cutting-edge model architectures, demonstrating the robust applicability and adaptability of our approach. Additionally, we introduce the CKA similarity index \cite{kornblith2019similarity} as an innovative analytical tool for a nuanced evaluation of our enhancements. The manuscript has been thoroughly revised to improve clarity and precision, with expanded analyses, visualizations, and results sections to provide a comprehensive and compelling exposition of our findings, clearly articulating the significant contributions of this research.

\section{Related Work}
\subsection{Deep Architecture}
ConvNets have dominated the field of computer vision for the past decade. Although their origins date back to the 1980s \cite{lecun1989backpropagation}, their true potential for large-scale image classification was not realized until the release of AlexNet \cite{krizhevsky2012imagenet}. Since then, various ConvNets such as VGG \cite{simonyan2014very}, ResNet \cite{he2016deep}, DenseNet \cite{huang2017densely}, Faster R-CNN \cite{ren2016faster}, FCN \cite{long2015fully}, Mask R-CNN \cite{long2015fully}, ResNeXt \cite{xie2017aggregated}, and EfficientNet \cite{tan2019efficientnet} have been proposed to enhance performance across different vision tasks \cite{qiao2023efficient, zhang2024partial}. In contrast, transformer architectures for vision tasks only began to emerge in the 2020s. The Vision Transformer (ViT) \cite{dosovitskiy2020image} demonstrated an impressive trade-off between speed and accuracy when applied to image classification using large training datasets like ImageNet-21K \cite{russakovsky2015imagenet} and JFT-300M \cite{sun2017revisiting}. Subsequent modifications to ViT aimed to improve architectural design to further enhance performance in vision tasks \cite{han2021transformer, yuan2021tokens, wang2021pyramid, liu2021swin, lin2022cat}. Despite the differences in architecture between ConvNets and transformers, some studies \cite{wang2018non, hu2019local, wu2021cvt, srinivas2021bottleneck, guo2022cmt} focus on combining convolutional and transformer architectures to leverage their respective advantages. For instance, while initially designed to enhance the performance of ConvNets through channel attention generation, recent studies \cite{chen2022mixformer} have shown that the Squeeze-and-Excitation (SE) module also benefits transformer architectures. Therefore, our work not only demonstrates that the proposed BAv2 module outperforms the SE module on ConvNets but also illustrates its potential to enhance the performance of existing state-of-the-art transformers.

\subsection{Attention Mechanism}
The attention mechanism is essential for emphasizing important features while suppressing less valuable ones, allowing the model to focus on significant regions within the context. As a component of the attention mechanism, channel attention has been widely utilized across various computer vision tasks, including image classification \cite{hu2018squeeze, zhang2024catnet, qing2024mpsa, li2024dual, lau2024large, luo2020learning, chen2024transform, ma2024coordinate} and downstream applications \cite{zhao2019pyramid, chen2019mixed, jang2019densenet, fu2019dual, lin2024mobilenetv2}. The pioneering work SENet \cite{hu2018squeeze} introduced an innovative mechanism for learning channel attention, achieving commendable performance. The SE module effectively recalibrates channel importance by aggregating spatial information with Global Average Pooling (GAP) and modeling inter-channel relationships through two fully connected layers.

Subsequent research aimed at enhancing SENet has primarily focused on refining channel feature representations and integrating channel attention with other attention mechanisms. For instance, CBAM \cite{woo2018cbam} demonstrated that reliance solely on GAP can lead to sub-optimal channel feature extraction, thus employing both GAP and Global Max Pooling (GMP) to aggregate features and prevent information loss. Similarly, SRM \cite{lee2019srm} and GSoPNet \cite{gao2019global} proposed global standard deviation pooling and global second-order pooling as alternatives to GAP, respectively. Stochastic region pooling \cite{luo2020stochastic} enhances the discrimination of feature maps across channels, thereby improving the representativeness and diversity of channel descriptors. Additionally, FcaNet \cite{qin2021fcanet} considers GAP as a special case of the Discrete Cosine Transform (DCT) and generalizes the aggregation of channel features in the frequency domain.

Moreover, numerous studies have explored the integration of channel attention with other attention mechanisms to produce more accurate attention maps. During channel attention extraction, methods like BAM \cite{park2018bam}, CBAM \cite{woo2018cbam}, GENet \cite{hu2018gather}, scSE \cite{roy2018recalibrating}, and TA \cite{misra2021rotate} leverage spatial information and employ convolutions to generate spatial attention. The Coordinate Attention (CA) \cite{hou2021coordinate} further extends this concept by decomposing spatial attention into two branches along horizontal and vertical directions.

Furthermore, self-attention \cite{vaswani2017attention} has gained traction in computer vision due to its capacity to capture long-range interactions, with notable examples including NLNet \cite{wang2018non}, $A^2$Net \cite{chen20182}, GCNet \cite{cao2019gcnet}, DANet \cite{fu2019dual}, and AANet \cite{bello2019attention}, all of which utilize various forms of self-attention to effectively extract global contextual information by modeling long-range dependencies. However, existing methods mainly focus on developing intricate attention modules within a single convolutional layer, seeking improved performance while overlooking the potential contributions from features of preceding layers. In contrast, our proposed BAv2 aims to investigate how prior features influence channel attention.

\subsection{Cross-layer Integration}
\begin{figure}[!t]
\centering
    \subfloat[]{\includegraphics[width=0.07\textwidth]{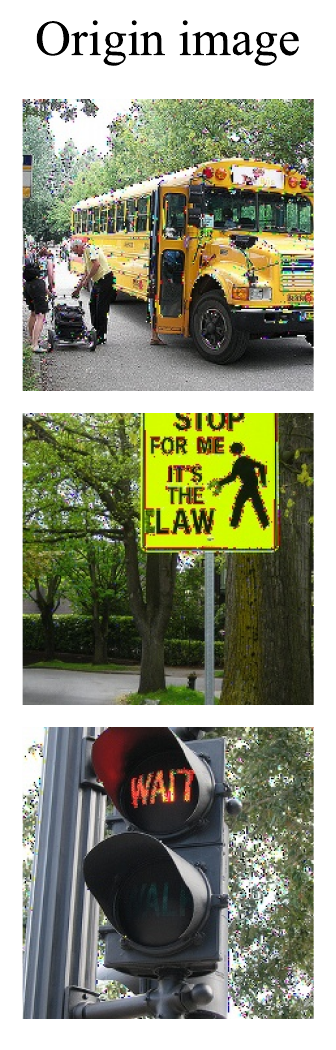}%
    \label{fig_2_a}}
    \subfloat[]{\includegraphics[width=0.07\textwidth]{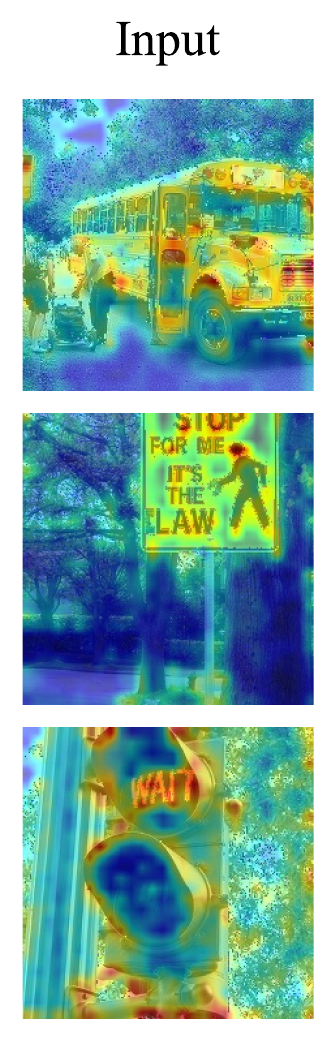}%
    \label{fig_2_b}}
    \subfloat[]{\includegraphics[width=0.07\textwidth]{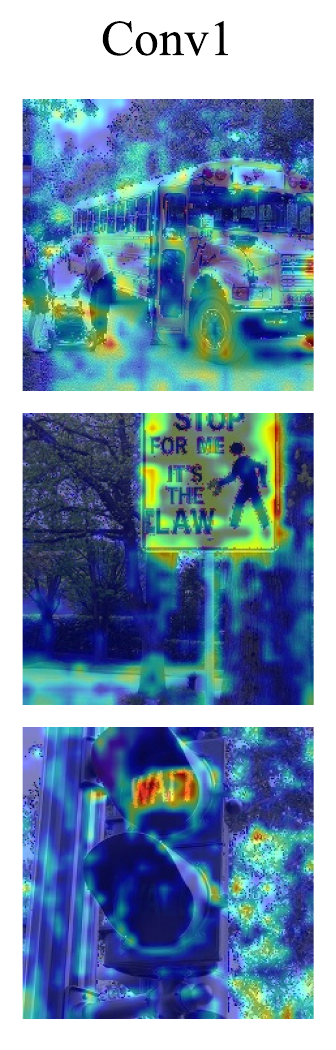}%
    \label{fig_2_c}}
    \subfloat[]{\includegraphics[width=0.07\textwidth]{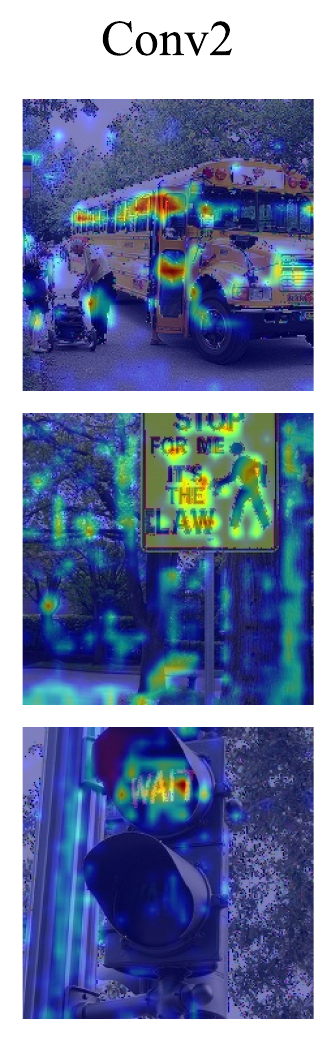}%
    \label{fig_2_d}}
    \subfloat[]{\includegraphics[width=0.07\textwidth]{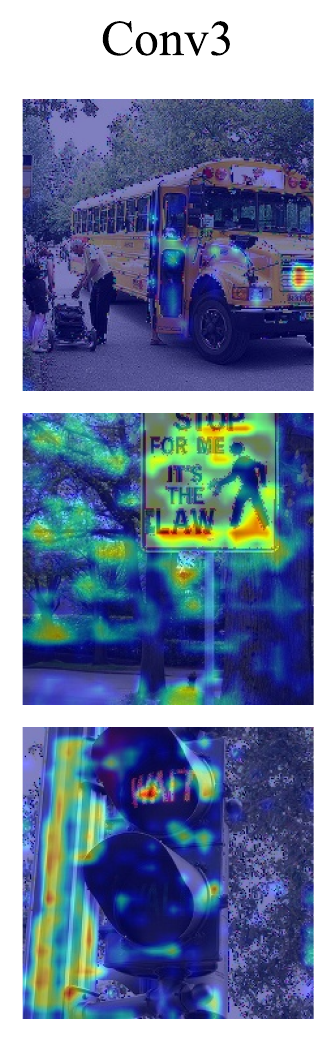}%
    \label{fig_2_e}}
    \subfloat[]{\includegraphics[width=0.07\textwidth]{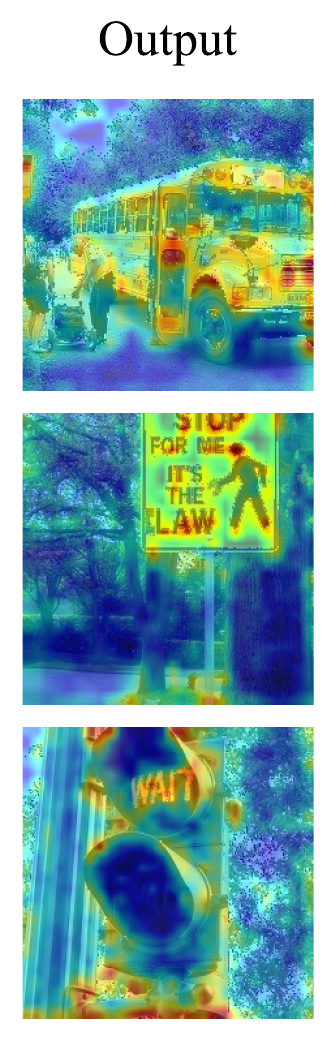}%
    \label{fig_2_f}}
\caption{Attention maps visualization for different convolutional layers in a Bottleneck block of ResNet50 \cite{he2016deep}. (a) Images from the validation set of ImageNet. (b) The grad-CAM \cite{selvaraju2017grad} visualization of the Bottleneck block input. (c)-(e) The grad-CAM visualization of the three convolutional layers output in the Bottleneck block. (f) The grad-CAM visualization of the Bottleneck block output.}
\label{fig_2}
\end{figure}

The utilization of skip connection is a commonly employed methodology in the domain of deep learning, which effectively alleviates gradient dispersion and enables the exploration of comprehensive features. It enhances the training of deep networks, making it feasible to use deeper network structures, and has gained widespread adoption in neural network design. The stacked convolutional layers in Fig. \ref{fig_2} face challenges like information loss and network degradation during transmission. To address this, ResNet \cite{he2016deep} introduces a residual connection that directly incorporates the input (Fig. \ref{fig_2}\subref{fig_2_a}) with the output of the last convolutional layer (Fig. \ref{fig_2}\subref{fig_2_e}). This ensures crucial information retention in the final output (Fig. \ref{fig_2}\subref{fig_2_f}) and effectively mitigates the vanishing gradient problem. Furthermore, Fig. \ref{fig_2} illustrates different convolutional layers focus on distinct feature regions while retaining relatively more feature information in previous layers. This observation inspires us to bridge outputs from previous convolutional layers for better attention map generation.

DenseNet \cite{huang2017densely} establishes dense connections between convolutional layers, significantly strengthening feature propagation and fusion. The U-Net family \cite{ronneberger2015u} utilizes skip connections to recover lost spatial information in the decoder. Additionally, the cross-layer techniques has been employed to enhance attention mechanisms efficacy further. DIANet \cite{huang2020dianet} proposed DIA unit, utilizing LSTM to share an attention module among different network layers, integrating hierarchical information and stabilizing deep network training robustly. Broad attention \cite{li2023bvit} introduces broad connections without discarding deep features, demonstrating the beneficial of connecting self-attention across different layers for enhancing vision transformers’ performance. In comparison with these works, our proposed BAv2 module aims to improve the SE module by bridging features from previous layers of the current block while introducing minimal parameters and computational overhead.

\section{Methodology}
To ensure a comprehensive description of the proposed bridge attention, we initially revisit the traditional channel mechanism (i.e., SENet \cite{hu2018squeeze}). Then, we empirically diagnose the limitations of generating attention maps solely at a single convolutional layer in existing channel attention mechanism. This serves as motivation for us to propose the Bridge Attention mechanism. Additionally, We provide a detailed description of our BAv2 module implementation and demonstrate its adoption for various deep architectures.

\subsection{Revisit SENet}
The channel attention mechanism explicitly establishes inter-dependencies among channels, thereby improving the model sensitivity towards the informative channels that have greater impact on determining the final classification result. Let $X = [x_1, x_2, \dots, x_c] \in \mathbb{R}^{C\times H\times W}$ represent the input of the SE module, where $C$ denotes the number of channels, $H$ represents the height of the feature, and $W$ indicates the width of feature. Consequently, we can express the attention weights generated within the SE module as follows:
\begin{equation}
\omega = \sigma (\mathcal{F_C}(gap(X))) \label{SE}
\end{equation}
where $gap(X)=\frac{1}{HW}\sum_{i=1,j=1}^{H,W}X_{i,j}$ represents channel-wise global average pooling, while $\sigma(\cdot)$ denotes the Sigmoid function. $\mathcal{F_C}(\cdot)$ signifies a concatenation of two stacked Full Connection (FC) layers, let $y=gap(x)$, which can be expressed as follows:
\begin{equation}
\mathcal{F_C}(y) = \bm{W_2} ReLU (\bm{W_1} y) \label{FC}
\end{equation}

In (\ref{FC}), ReLU represents the activation function. $\bm{W_1}$ and $\bm{W_2}$ are two linear transformation matrices used to adaptively recalibrate of channel relationships. The size of two matrices are set to $C \times (\frac{C}{r})$ and $(\frac{C}{r}) \times C$ respectively, where the reduction ratio $r$ is used to limit computation and complexity in the attention module.

We consider that SENet can be re-divided into two distinct components: \textbf{Integration} and \textbf{Generation}. In the SENet, the output $X$ undergoes global average pooling $gap(\cdot)$ to compress information across channels using matrix $\bm{W_1}$, which we define as \textbf{Integration} $\mathcal{I}(\cdot)$. Subsequently, the features are sequentially processed through a $RELU$ activation function, matrix $\bm{W_2}$ and $\sigma(\cdot)$ to obtain the final attention weights, which we refer to \textbf{Generation} $\mathcal{G}(\cdot)$. Therefore, we can express the general form of attention mechanism as follows:
\begin{align}
\mathcal{I}(\cdot) = \bm{W_1}(gap&(\cdot)),\  \mathcal{G}(\cdot) = \sigma(\bm{W_2}(ReLU(\cdot)))\\
&\omega = \mathcal{G}(\mathcal{I}(X)) \label{common form}
\end{align}

In our approach, we employ the $bridge$ operation in $\mathcal{I}(\cdot)$ to integrate more comprehensive features from preceding convolutional layers with the those of adjacent convolutional layer, resulting in enhanced attention generation.

\subsection{Limitation} \label{Limitation}
\begin{figure}[!t]
\centering
\includegraphics[width=0.45\textwidth]{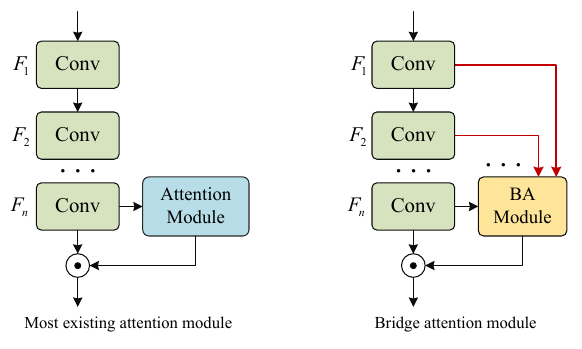}
\caption{The comparison between the prevalent attention module and our bridge attention module is presented in terms of their block structures. It can be observed that the BA module bridges the features from preceding convolutional layers, as indicated by teh red arrows.}
\label{fig_3}
\end{figure}

Fig. \ref{fig_3} illustrates the architecture commonly employed in existing attention methods within a basic block of ConvNets, comprising stacked convolutional layers and an attention module denoted as $\mathcal{F}(\cdot)$ and $att(\cdot)$ respectively. Therefore, the entire procedure can be depicted in the subsequent manner:
\begin{equation}
    \mathcal{F}_{att}(\cdot) = \mathcal{F}(\cdot) \odot att(\mathcal{F}(\cdot)) \label{Fatt}
\end{equation}
$\odot$ represents the element-wise multiplication.

We assume that the total number of stacked convolutional layers is represented by $n$, thus:
\begin{equation}
    \mathcal{F}(\cdot) = F_n(F_{n-1}(\cdots F_2(F_1))) \label{Fstack}
\end{equation}
$F_i(\cdot)$ represents the $i$-th convolutional layer within the block, where $1\leq i \leq n$.

Considering the distance, we assume that the outputs of $\mathcal{F}(\cdot)$ are more implicitly correlated with the previous $q$ convolutional layers,  thus (\ref{Fstack}) can be approximately equal to:
\begin{align}
    \mathcal{F}(\cdot) {} & \approx F_n(F_{n-1}(\cdots F_{n-(q-1)}(F_{n-q}))) \notag \\
                       {} & \Rightarrow \mathcal{F}(F_n, F_{n-1}, \dots, F_{n-(q-1)}, F_{n-q})
\end{align}

In most existing attention methods, the attention module only extracts features from the adjacent convolutional layer $F_n(\cdot)$. Some methods even have complicated calculations in $att(\cdot)$ to obtain richer information, which weakens its correlation with the previous $q$ convolutional layers:
\begin{align}
    att(\cdot) {} & = att(F_n(F_{n-1}(\cdots F_{n-(q-1)}(F_{n-q}))))  \notag \\
               {} & \approx att(F_n) \label{weaken}
\end{align}

The attention weights generated, as mentioned in (\ref{weaken}), exhibit a limited correlation with preceding convolutional layers, resulting in insufficient adaptability to the outputs of $\mathcal{F}(\cdot)$.

In fact, the aforementioned issue has been noticed by \cite{huang2020dianet, wang2021evolving}, however, only the attention weights from previous blocks are incorporated into the current attention module. In our method, we bridge features from the previous convolutional layers within the block are to the current attention module.

\subsection{Implementation of the BA module}
In this section, we provide a comprehensive account of how our method seamlessly integrates features from both preceding convolutional layers and adjacent convolutional layer within the block. Subsequently, we present a detailed explanation of the implementation details pertaining to Bridge Attention module.

\begin{figure*}[!t]
\centering
\includegraphics[width=0.9\textwidth]{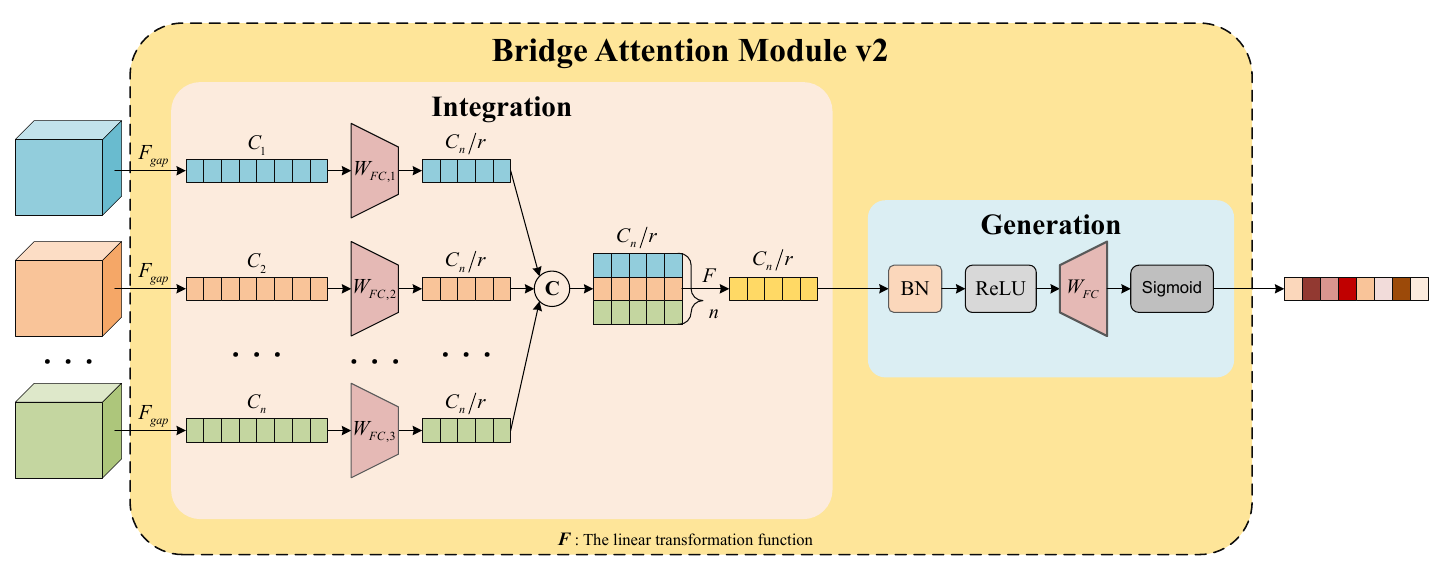}
\caption{The overview of Bridge Attention-v2 module is provided. BAv2 dynamically combines squeezed features obtained from various convolutional layers using the imporved bridge operation to generates channel weights.}
\label{fig_4}
\end{figure*}

Let the output of the $i$-th convolutional layer $F_i(\cdot)$ within the block be $X_i \in \mathbb{R}^{C_i\times H\times W}$. The first step is to generate a corresponding scalar representation for each channel of $X_i$. This process is considered as a compression problem \cite{qin2021fcanet}, and there are some methods \cite{woo2018cbam, lee2019srm, gao2019global, qin2021fcanet} that employ complicated strategies on adjacent convolutional layer to produce scalars that better represent their respective channels. However, as discussed in Sec. \ref{Limitation}, the scalars will lack correlation with the preceding layers. To mitigate this problem, we have chosen the simple and effective global average pooling as the compression method. We have empirically confirmed that channel representations generated using $gap(\cdot)$ are better at promoting information fusion of different convolutional layers when performing bridging (see Sec. \ref{Ablation}), showing the effectiveness of our design choice.

To effectively utilize the squeezed features, we introduce the bridge operation. The operation must meet three criteria. First, it should be simple and cheap to process the features of multiple convolutional layers. Second, it must ensure that the dimensions of corresponding features from each layer are the same in order to facilitate the subsequent fusion. Lastly, it is imperative for the model to effectively capture interdependencies among channels across diverse convolutional layers.

For better understanding, let $z_i \in \mathbb{R}^{C_i\times 1\times 1}$ be the output produced by $X_i$ after $gap(\cdot)$. Then we fed the $z_i$ respectively into corresponding matrices $\bm{W_{1,i}}$ of size $C_i \times (\frac{C_n}{r})$:
\begin{equation}
    S_i = \bm{W_{1,i}}(gap(X_i)) = \bm{W_{1,i}}(z_i)
\end{equation}
where $C_n$ is the number of channel in $X_n$ and $r$ is the reduction ratio same as in the SENet. Thus, each $S_i$ has the same size $\frac{C_n}{r}\times 1\times 1$.

\begin{figure}[!t]
\centering
\includegraphics[width=0.45\textwidth]{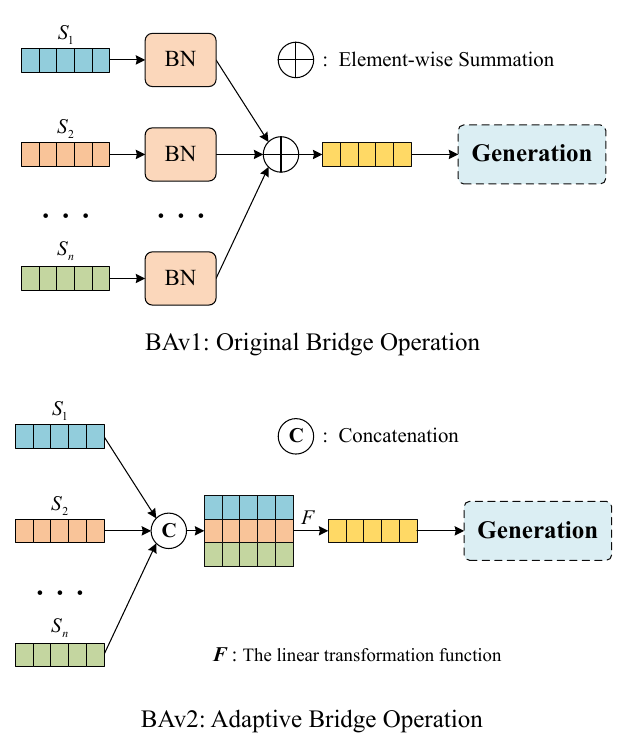}
\caption{The improved bridge operation in BAv2.}
\label{fig_5}
\end{figure}

For $S_i$ fusion, our BAv1 \cite{zhao2022ba} employs Batch Normalization \cite{ioffe2015batch} for each $S_i$, and then  add them directly by element-wise. In contrast, we propose an adaptive feature fusion approach in BAv2, as shown in Fig. \ref{fig_5}. This enables the BA module to dynamically integrate features from different convolutional layers during training, thereby mitigating redundancy and information interference. Specifically, we first concatenate all the features and then send them to a linear transformation function $F$ of size $n\times 1$, yielding
\begin{equation}
    S = F([S_1, S_2, \dots, S_n])
\end{equation}
where $n$ represents the total number of $S_i$, $[\cdot, \cdot]$ denotes the concatenation operation and $S$ is the output feature map of the \textbf{Integration} part.

Furthermore, in BAv1, BN is applied to each squeezed feature before integration in to ensure they have similar distributions. However, this approach is inefficient and may limit the diversity of representations. And the element-wise summation results in equivalent weight of each $S_i$ to the channel attention generation, whereas in fact, the contributions of each $S_i$ vary. Therefore, we employs adaptive feature fusion in BAv2 which enables the network to dynamically adjust the proportion of different features for optimal utilization across various convolutional layers. Nevertheless, BN can improve non-linear representation of the features and facilitate network parameter updates. Hence, we introduce a BN layer into the \textbf{Generation} part of BAv2. As a whole, the \textbf{Generation} part can be expressed as:
\begin{equation}
    \mathcal{G}_{BA}(\cdot) = \sigma(\bm{W_2}(ReLU(BN(S)))
\end{equation}
Then we input the integrated feature into \textbf{Generation} part and get the final attention weights. The overview of BAv2 module is shown in Fig. \ref{fig_4}.

Regarding the parameter calculation for the module, in the \textbf{Integration} part, BAv1 has $n \times \frac{C_n}{r}$ parameters due to BN for each $S_i$, where $n$ represents the total number of $S_i$. However, in BAv2, learnable parameters are only present when using the linear transformation function $F$, resulting in a total of $n$ parameters. In terms of the \textbf{Generation} part, BAv2 introduces an additional BN layer compared to BAv1, contributing $\frac{C_n}{r}$ extra parameters. To summarize, the parameter count for BAv2 changes from its original value of $n \times \frac{C_n}{r}$ to $n + \frac{C_n}{r}$.

\subsection{BA Module for ConvNet Architectures}
\begin{figure}[!t]
\centering
\includegraphics[width=0.45\textwidth]{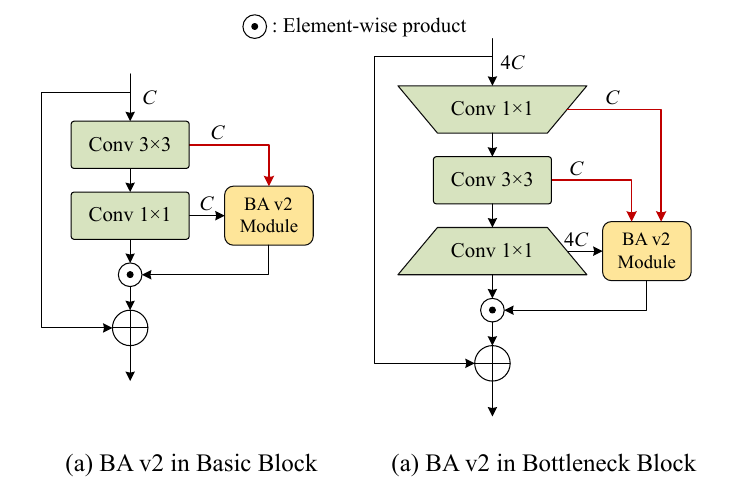}
\caption{Module Instantiation for ConvNets architectures. (a) BAv2 in Basic Block; (b) BAv2 in Bottleneck Block.}
\label{fig_6}
\end{figure}

The BAv1 module has proven that the attention module benefits from closer convolutional outputs, and by bridging all convolutional outputs of a block, improved performance can be achieved. Therefore, the BAv2 module can be directly applied to ConvNets by replacing the BAv1 module with it using the same configuration. For example, Fig. \ref{fig_6} shows how we integrated BAv2 module into the residual block in ResNet50. Moreover, we employ our approach to assess the capability of BAv2 module on stronger ConvNets, such as ResNeXt \cite{xie2017aggregated} and RegNetY \cite{radosavovic2020designing}. These networks have a similar architecture to ResNet, allowing us to easily integrate BAv2 modules into their blocks.

\subsection{BA Module for Vision Transformer Architectures}
\begin{figure}[!t]
\centering
\includegraphics[width=0.45\textwidth]{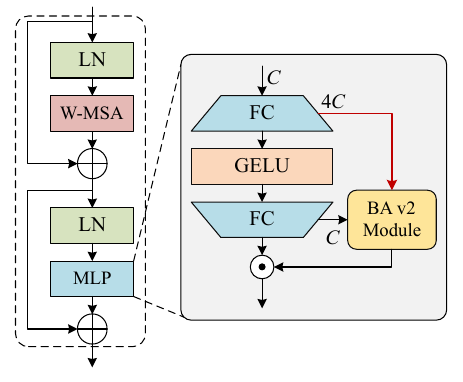}
\caption{Module instantiation for a Swin Transformer block \cite{liu2021swin}. The BAv2 module is integrated to the MLP layer.}
\label{fig_7}
\end{figure}

While channel attention modules have found extensive application in ConvNets, there remains a notable lack of research focusing on their integration with vision transformer architectures. One significant contribution in this area is the MixFormer \cite{chen2022mixformer}, which introduces a novel architecture that concurrently integrates local window self-attention and depthwise convolution. Additionally, it incorporates both channel and spatial attention to facilitate bi-directional interaction between these two branches. However, the design of MixFormer is fundamentally parallel, which limits its adaptability and transferability to other vision transformer frameworks. This gap in research highlights the need for more versatile approaches that can effectively merge channel attention mechanisms with various transformer architectures, potentially enhancing their performance and applicability across a broader range of tasks.

Our goal is to preserve the plug-and-play nature of the attention module, allowing for easy integration. The BAv2 module's flexibility enables its seamless application within vision transformers. In vision transformers, the feed-forward networks function as channel-mixing MLP layers \cite{tolstikhin2021mlp}. However, traditional MLP layers employ point-wise operations through two fully connected layers, which do not allow for cross-channel information to be effectively learned. To address this limitation, we position the BAv2 module before the residual connection of the MLP block, thereby enhancing feature extraction and fusion in the channel dimension.

As illustrated in Fig. \ref{fig_7}, the BAv2 module connects the outputs of the two fully connected layers and applies the generated channel attention weights to the output of the second fully connected layer. It is important to note that multiple strategies can be employed to integrate the BAv2 module into vision transformers. To assess the sensitivity of the insertion location of the BAv2 module, we conduct ablation experiments that explore various module designs, as detailed in Section \ref{Ablation}. This thorough evaluation ensures that we understand the optimal configuration for maximizing the performance of vision transformers.

\section{Experiments}
In this section, we evaluate the BAv2 on image classification tasks using ImageNet \cite{russakovsky2015imagenet} and CIFAR-10/100 \cite{krizhevsky2009learning} datasets at first, while comparing it to state-of-the-art attention-based methods. Subsequently, our approach is applied to various prominent deep neural network architectures, such as ResNeXt \cite{xie2017aggregated}, RegNet-Y \cite{radosavovic2020designing}, PVT v1 \cite{wang2021pyramid}, Swin Transformer \cite{liu2021swin} and CSWin-Transformer \cite{dong2022cswin} to verify the generality of the BAv2 module. Following that, a series of ablation analyses are conducted on the proposed BAv2 module. Lastly, we present a comparative evaluation of our proposed method against other attention-based approaches on object detection and instance segmentation tasks.

\begin{table*}
	\centering
	\caption{The default training hyper-parameters are employed in our method, unless explicitly specified.}
	\begin{tabular}{c | c | c | c | c | c | c | c | c | c | c | c | c | c}
	\toprule
        Epochs &Optimizer &\makecell[c]{Batch \\ size} &\makecell[c]{Learning \\ rate} &\makecell[c]{LR \\ decay} &\makecell[c]{Weight \\ decay} &\makecell[c]{Warmup \\ epochs} &\makecell[c]{Label \\ smooth} &\makecell[c]{Drop \\ path} &\makecell[c]{Repeated \\ Aug.} &\makecell[c]{Rand \\ Aug.} &Mixup &Cutmix &\makecell[c]{Erasing \\ Prob.}\\
	\midrule
        300 &AdamW &1024 &1e-3 &cosine &0.05 &20 &0.1 &0.1 &\checkmark &9/0.5 &0.8 &1.0 &0.25 \\
	\bottomrule
	\end{tabular}
	\label{setting}
\end{table*}

\begin{table*}
	\centering
	\caption{The performance of various attention methods on ImageNet is evaluated based on network parameters(Param.), floating point operations per second (FLOPs), and Top-1/Top-5 accuracy. The term $self.$ indicates that the metrics were obtained from our own retrained experiments, while $org.$ indicates that the metrics were reported in the original paper. The best result is highlighted in \textbf{\textcolor{blue}{bold with blue}}, whereas the second best result is indicated in \textcolor{green}{green}.}
	\begin{tabular}{l | c | c | c | c | c | c}
	\toprule
        Method &Years &Backbone &Parameters &FLOPs &\makecell[c]{Top-1 \\ \midrule $self. \ |\ \ org.$} &\makecell[c]{Top-5 \\ \midrule $self. \ |\ \ org.$}\\
	\midrule
        \makecell[l]{ResNet \cite{he2016deep} \\ + SE \cite{hu2018squeeze} \\ + FCA \cite{qin2021fcanet} \\ + BAv1(ours) \cite{zhao2022ba} \\ \textbf{+ BAv2(ours)}}
        &\makecell[c]{CVPR16 \\ CVPR18 \\ ICCV21 \\ ECCV22 \\ \textbf{---}}
        &\makecell[c]{ResNet-50}
        &\makecell[c]{25.56M \\ 28.07M \\ 28.07M \\ 28.71M \\ 28.70M}
        &\makecell[c]{4.13G \\ 4.14G \\ 4.14G \\ 4.15G \\ 4.15G}
        &\makecell[c]{
            78.88 $|$\ 75.20 \\ 
            80.06 $|$\ 76.71 \\ 
            80.07 $|$\ 78.57 \\
            \textcolor{green}{80.19} $|$\ 78.85 \\
            \textbf{\textcolor{blue}{80.49}}}
        &\makecell[c]{
            94.43 $|$\ 92.52 \\ 
            94.97 $|$\ 93.38 \\
            94.96 $|$\ 94.10 \\ 
            \textcolor{green}{95.05} $|$\ 94.28 \\ 
            \textbf{\textcolor{blue}{95.18}}}\\
	\midrule
        \makecell[l]{ResNet \cite{he2016deep} \\ + SE \cite{hu2018squeeze} \\ + FCA \cite{qin2021fcanet} \\ + BAv1(ours) \cite{zhao2022ba} \\ \textbf{+ BAv2(ours)}}
        &\makecell[c]{CVPR16 \\ CVPR18 \\ ICCV21 \\ ECCV22 \\ -}
        &\makecell[c]{ResNet-101}
        &\makecell[c]{44.55M \\ 49.29M \\ 49.29M \\ 50.49M \\ 50.48M}	
        &\makecell[c]{7.87G \\ 7.88G \\ 7.87G \\ 7.89G \\ 7.89G}
        &\makecell[c]{
            80.98 $|$\ 76.83 \\ 
            81.23 $|$\ 77.62 \\ 
            81.46 $|$\ 79.63 \\
            \textcolor{green}{81.51} $|$\ 80.03 \\ 
            \textbf{\textcolor{blue}{81.75}}}
        &\makecell[c]{
            95.27 $|$\ 93.48\\ 
            95.55 $|$\ 93.93\\ 
            95.63 $|$\ 94.63\\
            \textbf{\textcolor{blue}{95.77}} $|$\ 94.83 \,\\
            \textcolor{green}{95.66}}\\
	\bottomrule
	\end{tabular}
	\label{ImageNet}
\end{table*}

\subsection{Datasets and Experimental Settings}
\emph{1) Datasets:} 
Our proposed module’s performance has been validated through experiments conducted on ImageNet \cite{russakovsky2015imagenet}, which comprises 1.28 million training images and 50,000 validation images from 1,000 classes. Additionally, we evaluate the generalization capability of our BAv2 module by training it from scratch on smaller datasets such as CIFAR-10/100 \cite{krizhevsky2009learning}, each consisting of 60,000 $32\times32$ images belonging to 10 and 100 classes respectively. Both datasets are divided into 50,000 training images and 10,000 test images. To assess our method for object detection and instance segmentation tasks, we utilized the MS COCO2017 dataset \cite{lin2014microsoft} that includes 118K training images along with a validation set containing 5K images across 80 classes.

\emph{2) Implementation Details:} For the image classification task, we train all models from scratch using the training setting in DeiT \cite{touvron2021training} to fairly verify the competitiveness of our proposed approach in the current era. We employ a data augmentation approach that incorporates advanced techniques including random crop, random clip, Rand-Augment \cite{cubuk2020randaugment}, Random Erasing \cite{zhong2020random}, Mixup \cite{zhang2017mixup} and CutMix \cite{yun2019cutmix}. The training image resolution is fixed at $224\times224$ pixels for each model. Meanwhile, the validated images are initially resized to $256\times256$ pixels and subsequently cropped from the center to achieve a size of $224\times224$. Additionally, we employ Adamw \cite{loshchilov2017decoupled} optimizer for model training over a total of 300 epochs while incorporating a cosine decay learning rate scheduler. In addition to common settings, main advanced technologies include label smoothing \cite{szegedy2016rethinking}, DropPath \cite{larsson2016fractalnet}, and repeated augmentation \cite{hoffer2020augment}. Hyperparameters are provided in Table \ref{setting} for better comprehension.

For object detection and instance segmentation, we use Faster R-CNN \cite{ren2016faster} and Mask R-CNN \cite{he2017mask} as detectors, while the models pretrained on ImageNet serve as backbones. Our implementation of the detectors is based on the MMDetection toolkit \cite{chen2019mmdetection}, following its default configurations. To ensure consistency, we resize the input images to have a shorter side of 800 pixels while maintaining the longer side below 1,333 pixels. For optimizer, we utilize AdamW with a weight decay rate of 0.1 and set the batch size to 16. The training process lasts for a total of 12 epochs with an initial learning rate of 0.0001 that decreased by a factor of 10 at the 8th and 11th epoch correspondingly. All experiments are conducted on a workstation equipped with RTX 3090 GPUs.

\subsection{Image Classification on ImageNet}
We conducted a comparative analysis of our BAv2 module with several advanced attention methods, including SE \cite{hu2018squeeze}, FCA \cite{qin2021fcanet} and BAv1 \cite{zhao2022ba}, using ResNet50 and ResNet101 backbones on the ImageNet dataset. The evaluation criteria encompass both efficiency factors (i.e., network parameters, floating-point operations per second (FLOPs)) and effectiveness measures (i.e., Top-1/Top-5 accuracy). We present the metrics extracted from their original papers. It is worth mentioning that previous studies often suffer from inconsistencies because of outdated or suboptimal training strategies. Therefore, we have adopted the same training setup as Deit \cite{touvron2021training} to standardize our model training processes. This update allows for fairer and more accurate comparisons with prior attention methods and provides a contemporary benchmark for evaluating performance. The results are presented in Table \ref{ImageNet}, and our Github repository contains all these pre-trained models and code for reference. Under the same training setting, our BAv2 outperforms other attention methods, significantly outperforms SE by 0.43\% and 0.52\% in self. TOP-1 under the backbones of ResNet50 and ResNet101 respectively. Note that our reproduced results are significantly better than those of the original paper due to a longer training time and multiple data augmentation, with approximately 3\% improvement. Furthermore, BAv2 demonstrates a nontrival improvement of 0.30\% and 0.24\% under the two backbones when compared to BAv1 which already exhibited strong competitiveness.

\subsection{Image Classification on CIFAR10/100}
To investigate the generality of BAv2 on small datasets, we evaluate the proposed method using ResNet architecture on CIFAR10/100. The results show that our BAv2 consistently improves the TOP-1 accuracies across all baseline networks used in both datasets. Additionally, it performs competitively compared to other tested attention modules. On CIFAR10, our BAv2 module achieves classification accuracies of 97.22\%, 97.71\%, 97.79\% and 98.15\% based on ResNet18, ResNet34, ResNet50 and ResNet101 models respectively, outperforming all other methods as shown in Table \ref{CIFAR}.

\begin{table}
	\centering
	\caption{Performance comparisons are conducted among different attention methods applied to the CIFAR10/100 dataset, taking into account the numgher of parameters (Param.) and Top-1 accuracy. The best result is highlighted in \textbf{\textcolor{blue}{bold with blue}}.}
	\begin{tabular}{l | c | c}
	\toprule
    Method &\makecell[c]{CIFAR10\\ \midrule \, Param. $|$\ Top-1} &\makecell[c]{CIFAR100\\ \midrule \, Param.\ $|$\ Top-1}\\
	\midrule
        \makecell[l]{ResNet18 \cite{he2016deep} \\ + SE \cite{hu2018squeeze} \\ + FCA \cite{qin2021fcanet} \\ + BAv1(ours) \cite{zhao2022ba} \\ \textbf{+ BAv2(ours)}}
        &\makecell[c]{
            11.18M $|$\ 96.76 \\
            11.27M $|$\ 96.94 \\ 
            11.27M $|$\ 97.20 \\
            11.31M $|$\ 97.05 \\ 
            11.31M $|$\ \textbf{\textcolor{blue}{97.22}}}
        &\makecell[c]{
            11.23M $|$\ 82.40 \\ 
            11.31M $|$\ 82.59 \\ 
            11.31M $|$\ 82.56 \\ 
            11.36M $|$\ 82.69 \\ 
            11.36M $|$\ \textbf{\textcolor{blue}{82.90}}}\\
	\midrule
        \makecell[l]{ResNet34 \cite{he2016deep} \\ + SE \cite{hu2018squeeze} \\ + FCA \cite{qin2021fcanet} \\ + BAv1(ours) \cite{zhao2022ba} \\ \textbf{+ BAv2(ours)}}
        &\makecell[c]{
            21.29M $|$\ 97.16 \\ 
            21.45M $|$\ 97.42 \\ 
            21.45M $|$\ 97.61 \\ 
            21.53M $|$\ 97.50 \\ 
            21.53M $|$\ \textbf{\textcolor{blue}{97.71}}}
        &\makecell[c]{
            21.34M $|$\ 83.43 \\  
            21.49M $|$\ 83.61 \\ 
            21.49M $|$\ 83.87 \\ 
            21.57M $|$\ 83.90 \\  
            21.57M $|$\ \textbf{\textcolor{blue}{84.01}}}\\
    \midrule
        \makecell[l]{ResNet50 \cite{he2016deep} \\ + SE \cite{hu2018squeeze} \\ + FCA \cite{qin2021fcanet} \\ + BAv1(ours) \cite{zhao2022ba} \\ \textbf{+ BAv2(ours)}}
        &\makecell[c]{
            23.53M $|$\ 97.07 \\ 
            26.04M $|$\ 97.54 \\ 
            26.04M $|$\ 97.56 \\ 
            26.68M $|$\ 97.77 \\ 
            26.67M $|$\ \textbf{\textcolor{blue}{97.79}}}
        &\makecell[c]{
            23.71M $|$\ 83.47 \\ 
            26.23M $|$\ 84.75 \\ 
            26.23M $|$\ 84.35 \\ 
            26.86M $|$\ 85.22 \\  
            26.86M $|$\ \textbf{\textcolor{blue}{85.45}}}\\
    \midrule
        \makecell[l]{ResNet101 \cite{he2016deep} \\ + SE \cite{hu2018squeeze} \\ + FCA \cite{qin2021fcanet} \\ + BAv1(ours) \cite{zhao2022ba} \\ \textbf{+ BAv2(ours)}}
        &\makecell[c]{
            42.52M $|$\ 97.33 \\ 
            47.26M $|$\ 97.83 \\ 
            47.26M $|$\ 97.22 \\ 
            48.46M $|$\ 97.82 \\ 
            48.45M $|$\ \textbf{\textcolor{blue}{98.15}}}
        &\makecell[c]{
            42.71M $|$\ 84.40 \\ 
            47.45M $|$\ 85.30 \\ 
            47.45M $|$\ 84.50 \\  
            48.65M $|$\ 85.64 \\  
            48.64M $|$\ \textbf{\textcolor{blue}{85.98}}}\\
	\bottomrule
	\end{tabular}
	\label{CIFAR}
\end{table}

Similarly, on the CIFAR100 dataset, our BAv2 module achieves classification accuracies of 82.90\%, 84.01\%, 85.45\% and 85.98\% for the same base models, respectively. The BAv2 module, in the meantime, exhibits superior performance compared to their baseline models with enhancements of 0.50\%, 0.58\%, 1.98\% and 1.58\% for different base models, respectively. Furthermore, the BAv2 module with the ResNet50 model outperforms SE (84.75\%) by 0.70\%. Additionally, the ResNet101 model combined with the BAv2 module exhibits superior performance compared to the integration of the SE module with the ResNet101 model (85.30\%), surpassing it by a margin of 0.68\%. These results demonstrate that our BAv2 module exhibits excellent performance on small datasets and is not limited to specific datasets.

\subsection{Application on Other Backbones}
The bridge attention due to its versatile structural design, which can effectively improve various types of deep neural network architectures. We not only consider its application in ConvNets but also further explore the impact of the channel attention on the vision transformer architectures. We retrained all the original backbones while adding BAv2 module, and comparisons are shown in Table \ref{Diff_backbone}.

\begin{table}
	\centering
	\caption{A Performance comparison was conducted between advanced models and models incorporating the BAv2 module on the ImageNet dataset. In this study, $self.$ denotes metrics obtained from our own retraining experiments, while $org.$ refers to metrics reported in the original paper. The best result is highlighted in \textbf{\textcolor{blue}{bold with blue}}.}
	\begin{tabular}{l | c | c | c | c}
	\toprule
        Method &Years &Param. &FLOPs &\makecell[c]{Top-1\\ \midrule $self. \ |\ \ org.$} \\
	\midrule
        \makecell[l]{ResNeXt50 \cite{xie2017aggregated}\\ + BAv2\\ ResNeXt101 \cite{xie2017aggregated}\\ + BAv2}
        &\makecell[c]{CVPR17}
        &\makecell[c]{25.0M\\ 28.8M\\ 88.8M\\ 98.3M}
        &\makecell[c]{4.3G \\ 4.3G\\ 16.5G\\ 16.6G}
        &\makecell[c]{
            80.4 \ $|$\ \ 77.8\\ \textcolor{blue}{\textbf{81.2}(+0.8)} \\
            81.2 \ $|$\ \ 78.8\\ \textcolor{blue}{\textbf{81.6}(+0.4)}} \\
	\midrule
        \makecell[l]{RegNetY-4G \cite{radosavovic2020designing}\\ + BAv2}
        &\makecell[c]{CVPR20}
        &\makecell[c]{20.6M \\ 23.9M}
        &\makecell[c]{4.0G \\ 4.0G}
        &\makecell[c]{
            81.2 \ $|$\ \ 79.4\\ \textcolor{blue}{\textbf{81.5}(+0.3)}} \\
	\midrule
        \makecell[l]{Swin-T \cite{liu2021swin}\\ + BAv2\\ Swin-S \cite{liu2021swin}\\ + BAv2}
        &\makecell[c]{ICCV21}
        &\makecell[c]{28.8M\\ 29.1M\\ 50.0M\\ 51.0M}
        &\makecell[c]{4.4G\\ 4.4G\\ 8.6G\\ 8.6G}
        &\makecell[c]{
            81.2 \ $|$\ \ 81.3\\ \textcolor{blue}{\textbf{81.8}(+0.4)} \\
            \textbf{---}  \ $|$ \ 83.0\\ \textcolor{blue}{\textbf{83.2}(+0.2)}} \\
	\midrule
        \makecell[l]{PVTv1-S \cite{wang2021pyramid}\\ + BAv2\\ PVTv1-M \cite{wang2021pyramid}\\ + BAv2}
        &\makecell[c]{ICCV21}
        &\makecell[c]{24.5M\\ 24.7M\\ 44.2M\\ 44.9M}
        &\makecell[c]{3.7G\\ 3.7G\\ 6.5G\\ 6.5G}
        &\makecell[c]{
            \textbf{---} \ $|$\ \ 79.8\\ \textcolor{blue}{\textbf{80.4}(+0.6)} \\
            \textbf{---} \ $|$\ \ 81.2\\ \textcolor{blue}{\textbf{82.1}(+0.9)}} \\
    \midrule
        \makecell[l]{CSWin-T \cite{dong2022cswin}\\ + BAv2\\}
        &\makecell[c]{CVPR22}
        &\makecell[c]{23.0M\\ 23.0M}
        &\makecell[c]{4.3G\\ 4.3G}
        &\makecell[c]{
            82.7 \ $|$\ 82.8\\ \textcolor{blue}{\textbf{82.9}(+0.2)}}\\
	\bottomrule
	\end{tabular}
	\label{Diff_backbone}
    \begin{tablenotes}
        \footnotesize
        \item[1] Note: The performance comparison of YOLOv8(2023), YOLOv9(2024), and the model with BAv2 on COCO2017 is shown in Table \ref{YOLO_detection}
    \end{tablenotes}
\end{table}

For ResNeXt50 \cite{xie2017aggregated}, our reproduced result achieves a TOP-1 80.4\%, which is 2.6\% higher than the results reported in the original paper. By introducing BAv2 into ResNeXt50, we achieve a further improvement with a TOP-1 accuracy of 81.2\%, representing an increase of 0.8\% compared to our reproduced result. Similarly, by applying BAv2 to ResNeXt101, we achieve an even higher TOP-1 accuracy of 81.6\%. This is 0.4\% higher than our reproduced result (81.1\%) and shows an improvement of 2.8\% compared to the original paper. Furthermore, when suing BAv2 with RegNetY-4G \cite{radosavovic2020designing}, we observe a significant improvement of 0.3\% compared to our reproduced result and a gain of 2.1\% compared to the original paper. It is noteworthy that RegNetY is an extension of RegNetX with the inclusion of an SE module. Therefore, this comparison serves as a performance evaluation between the BAv2 module and the SE module.

Table \ref{Diff_backbone} is apparent that BAv2 demonstrates a performance improvement of 0.6\% and 0.9\% in comparison to the original PVT v1-S/M \cite{wang2021pyramid}. For Swin \cite{liu2021swin}, our BAv2 can still boost performance by 0.4\% on Swin-T and 0.2\% on Swin-S respectively. Even when facing CSWin-T \cite{dong2022cswin} with an 82.7\% Top-1 accuracy, BAv2 can still bring a 0.2\% performance improvement.

The experimental results presented above provide comprehensive evidence for the versatility of our BAv2 and its potential to enhance the performance of various deep neural network architectures. In summary, bridge attention can be universally introduced to ConvNets as a generic mechanism. The exceptional performance of BAv2 in vision transformer architectures further validates the effectiveness of channel attention for vision transformers and emphasizes the importance of BAv2 in leverage significant features.

\subsection{Ablation Studies} \label{Ablation}
\emph{1) Different pooling methods:} Given the availability of diverse pooling methods for generating channel representations, our aim is to assess their efficacy in order to identify the most suitable pooling method for BAv2. We compare four pooling methods: global average pooling \cite{hu2018squeeze}, joint use of average pooling and max pooling \cite{woo2018cbam}, discrete cosine transform \cite{qin2021fcanet}, and style pooling \cite{lee2019srm}. We evaluate ImageNet using ResNet50 as the backbone and present the corresponding results in Table \ref{Diff_pool}. However, when using only global average pooling, we achieve the best results. Experimental results indicate that complex strategies generate channel representation scalars lacking correlation with previous convolutional layers, resulting in information redundancy and interference that can affect performance. On the other hand, features generated by simple and direct global average pooling contribute more effective information to the attention module.

\begin{table}[!t]
	\centering
	\caption{Effect of using different pooling methods in BAv2-ResNet50 on ImageNet.}
	\begin{tabular}{l | c | c}
	\toprule
        Pooling mehthod &Top-1 &Top-5 \\
    \midrule
        ResNet50 &78.88 &94.43 \\
	\midrule
        \makecell[l]{AvgPool\\ AvgPool \& MaxPool\\ AvgPool \& StdPool\\ Discrete cosine transform}
        &\makecell[c]{\textbf{80.49}\\ 80.23\\ 80.31\\ 80.12}
        &\makecell[c]{\textbf{95.18}\\ 95.09\\ 95.07\\ 95.11} \\
	\bottomrule
	\end{tabular}
	\label{Diff_pool}
\end{table}

\begin{figure*}[!t]
\centering
\includegraphics[width=0.9\textwidth]{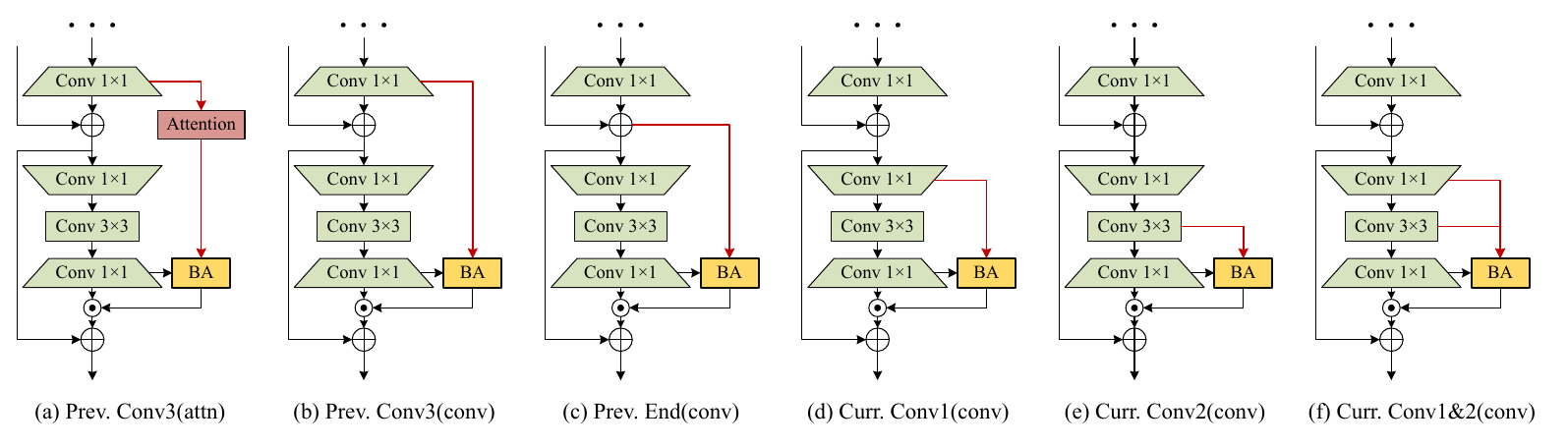}
\caption{The effectiveness of various previous features explored in the ablation study.}
\label{fig_8}
\end{figure*}

\emph{2) Different previous features:}
We explore the potential of integrating various preceding features with the attention module to assess their efficacy on channel attention. To expedite the experiments, we employ ResNet50 as the backbone, implementing identical data augmentation techniques and hyper-parameter configurations as mentioned in \cite{qin2021fcanet}, to evaluate its performance on ImageNet. The module structures of BAv1 bridge different features, as illustrated in Fig. \ref{fig_8}, and the corresponding results are presented in Table \ref{Diff_features}. Despite being at the same position, $row_3$ achieves higher accuracy than $row_2$. This is because the attention layer derives less benefit from incorporating attention weights in rescaling feature maps compared to convolutional outputs. Furthermore, achieving enhanced performance necessitates bridging proximate features as it enables substantial compression of features within the attention module, thereby facilitating the extraction of more relevant information from proximate features.

\begin{table}[!t]
    \centering
    \caption{The comparison involves the integration of diverse attributes, including attention weights ($attn$), outputs of convolutional layer ($conv$), preceding blocks ($prev.$), current blocks ($curr.$) adn the $i$-th convolutional layers of convolutional, as well as the end of the block ($end$).}
    \begin{tabular}{l | c | l | c | c  }
    \toprule
        Backbone &Type &Position &Param. &TOP-1 \\
    \midrule
        SE-ResNet50 &$None$ &\textbf{---} &28.07M &78.14 \\
    \midrule
        \makecell[l]{BAv1-ResNet50}
        &\makecell[c]{$attn$\\ $conv$\\ $conv$\\ $conv$\\ $conv$\\ $conv$}
        &\makecell[l]{$prev., conv_3$\\ $prev., conv_3$\\ $prev., end$\\ $curr., conv_1$\\ $curr., conv_2$\\ $curr., conv_{1\&2}$}
        &\makecell[c]{29.16M\\ 29.16M\\ 29.16M\\ 28.39M\\ 28.39M\\ 28.71M}
        &\makecell[c]{78.41\\ 78.49\\ 78.54\\ 78.78\\ 78.77\\ \textbf{78.85}} \\
    \bottomrule
    \end{tabular}
    \label{Diff_features}
\end{table}

\begin{figure}[!t]
\centering
\includegraphics[width=0.3\textwidth]{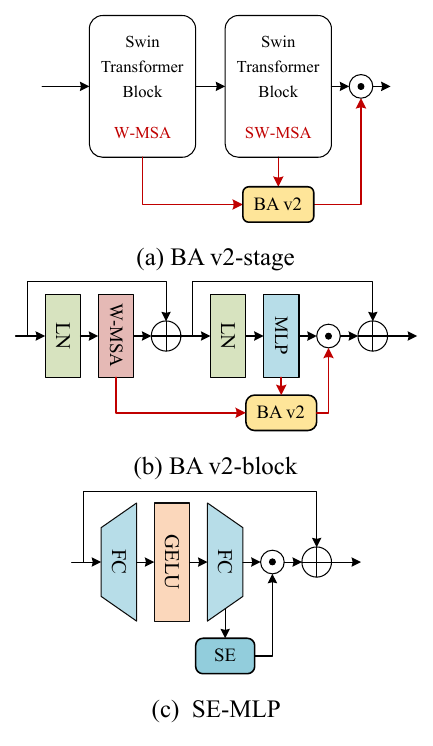}
\caption{Structural details of the three models in ablation study. (a) BAv2-stage, (b) BAv2-block, (c) SE-MLP.}
\label{fig_9}
\end{figure}

\emph{3) Different integration strategy:} 
Finally, an ablation analysis was employed to assess the impact of integrating the BAv2 module at different positions in vision transformer architectures. We train Swin-T as the backbone and assess its the performance on ImageNet. In addition to the proposed BAv2 design, we consider three variants: (1) BAv2-stage module, where the BAv2 module is placed within a stage to bridge the output of two adjacent transformer blocks; (2) BAv2-block module, where the BAv2 module is positioned inside a transformer block to bridge the output of the self-attention layer and the MLP layer. Furthermore, we also compare our BAv2 design with moving the SE module to the second FC layer in MLP, referred to as SE-MLP. The structure and performance of each variant is reported in Fig. \ref{fig_9} and Table \ref{Diff_integration}. It can be observed that the models with the attention module exhibit superior performance compared to the original Swin-T. The performance of BAv2-stage is better than BAv2-block and comparable to SE-MLP, but inferior to BAv2-MLP. This indicates a weaker correlation between the self-attention layer and the MLP layer in comparison to the correlation between two transformer blocks, while the strongest correlation is observed between two FC layers within the MLP layer. Moreover, compared with the SE module, our BAv2 demonstrates more advantages in vision transformer architectures.

\begin{table}[!t]
	\centering
	\caption{Effect of different BAv2 module integration strategies with Swin-T on ImageNet.}
	\begin{tabular}{l | c | c}
	\toprule
         Design &Param. &Top-1 \\
	\midrule
        \makecell[l]{Original Swin-T\\ BAv2-stage\\ BAv2-block\\ SE-MLP\\ BA-MLP}
        &\makecell[c]{28.3M\\ 28.5M\\ 28.7M\\ 28.5M\\ 29.1M}
        &\makecell[c]{81.16\\ 81.55\\ 81.45\\ 81.55\\ \textbf{81.77}} \\
	\bottomrule
	\end{tabular}
	\label{Diff_integration}
\end{table}

\subsection{Application}
In this subsection, we initially assess the performance of BAv2 on object detection tasks using classical Faster R-CNN \cite{ren2016faster} and Mask R-CNN \cite{he2017mask}. Furthermore, we present the results of instance segmentation utilizing the Mask R-CNN backbone. To facilitate a comparative analysis between BAv2 module and five other attention modules (SE \cite{hu2018squeeze}, ECA \cite{wang2020eca}, FCA \cite{qin2021fcanet}, SimAM \cite{yang2021simam} and BAv1 \cite{zhao2022ba}), all these attention modules can be applied in the detectors. The models undergo pretraining on ImageNet prior to fine-tuning on MS COCO2017. Finally, we explore the application of the BAv2 module on YOLOv8 \cite{Jocher_Ultralytics_YOLO_2023} and YOLOv9 \cite{wang2024yolov9} series models to verify its effectiveness in cutting-edge and state-of-the-art detection models.

The first experiment follows the majority of previous studies and report results in terms of $mAP$, $AP_{50}$ and $AP_{75}$. Table \ref{COCO_detection} presents the results obtained by different models on the MS COCO2017 validation set. According to Table \ref{COCO_detection}, the BAv2 module outperforms the referenced attention modules with both base models across all three evaluation metrics. Specifically, when using Faster R-CNN, BAv2 achieves a higher mAP than BAv1 by 1.6\% and 1.6\% on ResNet50 and ResNet101 respectively. If Mask R-CNN is used as the base detector instead, the gains increase by 1.2\%.

\begin{table}[!t]
    \centering
    \caption{Performance comparisons of different attention methods for object detection are conducted, with Average Precision ($AP$) serving as the primary evaluation metric. The best result is highlighted in \textbf{\textcolor{blue}{bold with blue}}. The second best result is highlighted in \textcolor{green}{green}.}
    \begin{tabular}{l | c | c | c | c | c}
    \toprule
        Backbone &Detector &Param. &$mAP$ &$AP_{50}$ &$AP_{75}$ \\
    \midrule
        \makecell[l]{ResNet50 \cite{he2016deep}\\ + SE \cite{hu2018squeeze}\\ + ECA \cite{wang2020eca}\\ + FCA \cite{qin2021fcanet}\\ + SimAM \cite{yang2021simam}\\ + BAv1 \cite{zhao2022ba}\\ \textbf{+ BAv2}\\ \midrule ResNet101 \cite{he2016deep}\\ + SE \cite{hu2018squeeze}\\ + ECA \cite{wang2020eca}\\ + FCA \cite{qin2021fcanet}\\ + SimAM \cite{yang2021simam}\\ + BAv1 \cite{zhao2022ba}\\ \textbf{+ BAv2}}
        &\makecell[c]{Faster-RCNN}
        &\makecell[c]{42M\\ 44M\\ 42M\\ 44M\\ 42M\\ 45M\\ 45M\\ \midrule 61M\\ 65M\\ 61M\\ 65M\\ 61M\\ 66M\\ 66M}
        &\makecell[c]{36.4\\ 37.7\\ 38.0\\ 39.0\\ 39.2\\ \textcolor{green}{39.5}\\ \textbf{\textcolor{blue}{41.1}}\\ \midrule 38.7\\ 39.6\\ 40.3\\ 41.2\\ 41.2\\ \textcolor{green}{41.7}\\ \textbf{\textcolor{blue}{43.3}}}
        &\makecell[c]{58.2\\ 60.1\\ 60.6\\ 61.1\\ 60.7\\ \textcolor{green}{61.3}\\ \textbf{\textcolor{blue}{63.0}}\\ \midrule 60.6\\ 62.0\\ 62.9\\ 63.3\\ 62.4\\ \textcolor{green}{63.4}\\ \textbf{\textcolor{blue}{65.0}}}
        &\makecell[c]{39.2\\ 40.9\\ 40.9\\ 42.3\\ 42.8\\ \textcolor{green}{43.0}\\ \textbf{\textcolor{blue}{44.7}}\\ \midrule 41.9\\ 43.1\\ 44.0\\ 44.6\\ 45.0\\ \textcolor{green}{45.1}\\ \textbf{\textcolor{blue}{47.4}}} \\
    \midrule
        \makecell[l]{ResNet50 \cite{he2016deep}\\ + SE \cite{hu2018squeeze}\\ + ECA \cite{wang2020eca}\\ + FCA \cite{qin2021fcanet}\\ + SimAM \cite{yang2021simam}\\ + BAv1 \cite{zhao2022ba}\\ \textbf{+ BAv2}}
        &\makecell[c]{Mask-RCNN}
        &\makecell[c]{44M\\ 47M\\ 44M\\ 47M\\ 44M\\ 47M\\ 47M}
        &\makecell[c]{37.2\\ 38.4\\ 39.0\\ 40.3\\ 39.8\\ \textcolor{green}{40.5}\\ \textbf{\textcolor{blue}{41.7}}}
        &\makecell[c]{58.9\\ 60.9\\ 61.3\\ \textcolor{green}{62.0}\\ 61.0\\ 61.7\\ \textbf{\textcolor{blue}{63.4}}}
        &\makecell[c]{40.3\\ 42.1\\ 42.1\\ 44.1\\ 43.4\\ \textcolor{green}{44.2}\\ \textbf{\textcolor{blue}{45.5}}} \\
    \bottomrule
    \end{tabular}
    \label{COCO_detection}
\end{table}

Then we used Mask R-CNN to evaluate our method on the instance segmentation task. The results are shown in Table \ref{COCO_instance}. BAv2 achieved a $mAP$ of 38.4\% and 40.0\% under two backbones, outperforming other attention modules by a significant margin. Specifically, BAv2 notably outperforms BAv1 by 1.8\% and 1.9\%, respectively, in terms of $mAP$. Consequently, BAv2 demonstrates exceptional performance not only in the realm of image classification but also in tasks pertaining to object detection and instance segmentation, thereby showcasing its good capacity for generalization across diverse computer vision assignments.

\begin{table}[!t]
    \centering
    \caption{Comparisons of performance across various attention methods for the instance segmentation are presented. The best result is highlighted in \textbf{\textcolor{blue}{bold with blue}}. The second best result is highlighted in \textcolor{green}{green}.}
    \begin{tabular}{l | c | c | c | c | c | c }
    \toprule
        Backbone &$mAP$ &$AP_{50}$ &$AP_{75}$ &$AP_S$ &$AP_M$ &$AP_L$\\
    \midrule
        \makecell[l]{ResNet50 \cite{he2016deep}\\ + SE \cite{hu2018squeeze}\\ + ECA \cite{wang2020eca}\\ + FCA \cite{qin2021fcanet}\\ + SimAM \cite{yang2021simam}\\ + BAv1 \cite{zhao2022ba}\\ \textbf{+ BAv2}\\ \midrule ResNet101 \cite{he2016deep}\\ + SE \cite{hu2018squeeze}\\ + ECA \cite{wang2020eca}\\ + SimAM \cite{yang2021simam}\\ + BAv1 \cite{zhao2022ba}\\ \textbf{+ BAv2}}
        &\makecell[c]{34.2\\ 35.4\\ 35.6\\ 36.2\\ 36.0\\ \textcolor{green}{36.6}\\ \textbf{\textcolor{blue}{38.4}}\\ \midrule 35.9\\ 36.8\\ 37.4\\ 37.6\\ \textcolor{green}{38.1}\\ \textbf{\textcolor{blue}{40.0}}}
        &\makecell[c]{55.9\\ 57.4\\ 58.1\\ 58.6\\ 57.9\\ \textcolor{green}{58.7}\\ \textbf{\textcolor{blue}{60.5}}\\ \midrule 57.7\\ 59.3\\ 59.9\\ 59.5\\ \textcolor{green}{60.6}\\ \textbf{\textcolor{blue}{62.6}}}
        &\makecell[c]{36.2\\ 37.8\\ 37.7\\ 38.1\\ 38.2\\ \textcolor{green}{38.6}\\ \textbf{\textcolor{blue}{41.1}}\\ \midrule 38.4\\ 39.2\\ 39.8\\ 40.1\\ \textcolor{green}{40.4}\\ \textbf{\textcolor{blue}{42.9}}}
        &\makecell[c]{16.1\\ 17.1\\ 17.6\\ \textbf{---}\\ \textbf{\textcolor{blue}{19.1}}\\ 18.2\\ \textcolor{green}{18.9}\\ \midrule 16.8\\ 17.2\\ 18.1\\ \textbf{\textcolor{blue}{20.5}}\\ 18.7\\ \textcolor{green}{20.2}}
        &\makecell[c]{37.5\\ 38.6\\ 39.0\\ \textbf{---}\\ \textcolor{green}{39.7}\\ 39.6\\ \textbf{\textcolor{blue}{41.8}}\\ \midrule 39.7\\ 40.3\\ 41.1\\ \textcolor{green}{41.5}\\ \textcolor{green}{41.5}\\ \textbf{\textcolor{blue}{43.3}}}
        &\makecell[c]{46.3\\ 51.8\\ 51.8\\ \textbf{---}\\ 48.6\\ \textcolor{green}{52.3}\\ \textbf{\textcolor{blue}{55.5}}\\ \midrule 49.7\\ 53.6\\ 54.1\\ 50.8\\ \textcolor{green}{54.8}\\ \textbf{\textcolor{blue}{58.0}}\\} \\
    \bottomrule
    \end{tabular}
    \label{COCO_instance}
\end{table}

To further evaluate the effectiveness of BAv2 in cutting-edge detection models, we incorporated the bridge attention mechanism into the high-performing YOLOv8 and YOLOv9 architectures. For our experiments on the MS COCO2017 detection task, we utilized YOLOv8s, YOLOv8m, GELANs, and GELANm as backbone networks. Notably, the GELAN model, introduced in YOLOv9 \cite{wang2024yolov9}, represents an improved neural network architecture called the Generalized Efficient Layer Aggregation Network. As presented in Table \ref{YOLO_detection}, the experimental setups and models were aligned with the official codebases of both Ultralytics \cite{Jocher_Ultralytics_YOLO_2023} and YOLOv9 \cite{wang2024yolov9}. We also reproduced the original models to ensure a fair comparison.

The results demonstrate that incorporating the BAv2 module led to an increase in $mAP$ of 0.5\% and $AP_{50}$ improvements of 0.6\% and 0.5\% for YOLOv8s and YOLOv8m, respectively. In comparison to GELANs and GELANm, $mAP$ improved by 0.6\% and 0.4\%, while $AP_{50}$ increased by 0.8\% and 0.5\%. These findings highlight the versatility and effectiveness of bridge attention, showcasing its ability to integrate seamlessly into various models as a general mechanism. Moreover, it brings substantial performance gains with a negligible increase in parameters.

\begin{table}[!t]
    \centering
    \caption{Performance comparison of YOLOv8, YOLOv9, and the models with BAv2 module on MS COCO2017. The best result is highlighted in \textbf{\textcolor{blue}{bold with blue}}}
    \begin{tabular}{l | c | c | c | c | c}
    \toprule
        Model &Size &Param. &FLOPs &$mAP$ &$AP_{50}$ \\
    \midrule
        \makecell[l]{YOLOv8s \\ \textbf{+ BAv2} \\ YOLOv8m \\ \textbf{+ BAv2}}
        &\makecell[c]{640\\ 640\\ 640\\ 640}
        &\makecell[c]{11.2M\\ 11.6M\\ 25.9M\\ 26.6M}
        &\makecell[c]{28.6G\\ 29.1G\\ 79.3G\\ 79.9G}
        &\makecell[c]{44.7\\ \textcolor{blue}{\textbf{45.2}(+0.5)}\\ 49.8\\ \textcolor{blue}{\textbf{50.3}(+0.5)}}
        &\makecell[c]{61.2\\ \textcolor{blue}{\textbf{61.8}(+0.6)}\\ 66.5\\ \textcolor{blue}{\textbf{67.0}(+0.5)}} \\
    \midrule
        \makecell[l]{GELANs \\ \textbf{+ BAv2} \\ GELANm \\ \textbf{+ BAv2}}
        &\makecell[c]{640\\ 640\\ 640\\ 640}
        &\makecell[c]{7.1M\\ 7.3M\\ 20.0M\\ 20.6M}
        &\makecell[c]{26.4G\\ 26.9G\\ 76.3G\\ 77.1G}
        &\makecell[c]{45.9\\ \textcolor{blue}{\textbf{46.5}(+0.6)}\\ 50.2\\ \textcolor{blue}{\textbf{50.6}(+0.4)}}
        &\makecell[c]{61.9\\ \textcolor{blue}{\textbf{62.7}(+0.8)}\\ 66.8\\ \textcolor{blue}{\textbf{67.3}(+0.5)}} \\
    \bottomrule
    \end{tabular}
    \label{YOLO_detection}
\end{table}

\subsection{Importance of the Integrated Features}
\begin{figure}[!t]
\centering
\includegraphics[width=0.45\textwidth]{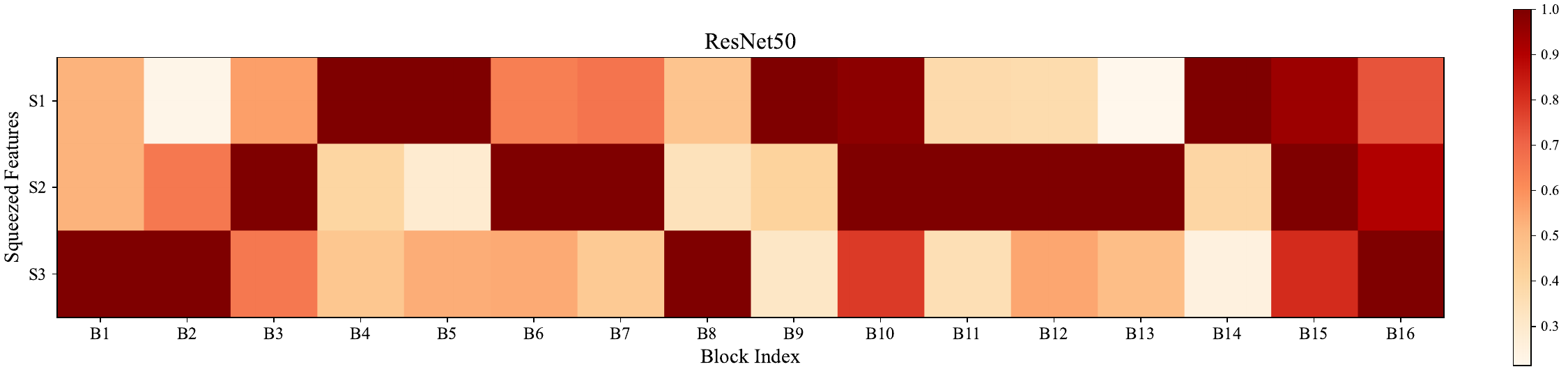}
\caption{Comparison of similarity is conducted using CKA to analyze the integrated features. In the BAv2-ResNet50 model, $B_i$ refers to the $i$-th block, while $S_i$ represents the $i$-th squeezed feature within a specific block. The color depth of each square in the comparison matrix indicates its level of similarity, reflecting its significance.}
\label{fig_10}
\end{figure}

In our approach, we integrate features from various convolutional layers within the block into the attention module. Our goal is to investigate the correlation between these integrated features and the generated attention weights, as well as to understand their respective contributions to the determination of these weights. To elucidate this relationship, we propose the use of Central Kernel Alignment (CKA) \cite{kornblith2019similarity}, a method that enables a quantitative assessment of feature similarity within models. By employing CKA, we can effectively measure the importance of each branch in the BA module, considering the similarity between the different integrated features within a block and the corresponding attention weights. This analysis not only provides insights into how each feature influences attention but also enhances our understanding of the underlying mechanisms driving the performance of the attention module. Specifically, given an ensemble feature $\textbf{X}$ and attention weights $\textbf{Y}$ as input, we can derive the Gram matrices $\textbf{K}=\textbf{X}\textbf{X}^T$ and $\textbf{L}=\textbf{Y}\textbf{Y}^T$. Then CKA can be computed as
\begin{equation}
    CKA(\textbf{K},\textbf{L}) = \frac{HSIC(\textbf{K}, \textbf{L})}{\sqrt{HSIC(\textbf{K}, \textbf{K})HSIC(\textbf{L}, \textbf{L})}}
\end{equation}
where HSIC is the Hilbert-Schmidt independence criterion \cite{gretton2007kernel}. The BAv2-ResNet50 model comprises 16 blocks, each consisting of three convolutional layers followed by an attention module. We validate the model using the ImageNet validation set, which allows us to obtain the squeezed feature $S_i$ and attention weights $\omega_i$ for each block. Subsequently, we compute the Central Kernel Alignment (CKA) similarity for $S_i$ and $\omega_i$ in each block, and we visualize the feature importance, as illustrated in Fig. \ref{fig_10}. The results reveal that, for most blocks, the features derived from earlier convolutional layers are more influential than $S_3$. In certain blocks, such as B10, B15, and B16, the three features contribute approximately equally to the attention weight. These findings indicate that the attention weights are significantly shaped by the features extracted from preceding convolutional layers, highlighting their critical role in shaping the amalgamated features across multiple blocks. This analysis underscores the importance of understanding feature interactions in enhancing the model's performance.

\section{Conclusion}
While our advancements in vision transformer architectures mark an initial step, future research will systematically explore how different combinations of transformer block outputs influence overall performance. Our objective is to develop a novel and robust channel attention module that significantly enhances the efficacy of transformer architectures.

In this paper, we introduced an enhanced bridge attention model, BAv2, which effectively addresses the limitations of traditional channel attention mechanisms that heavily rely on adjacent convolutional layers. By integrating features from multiple layers for a more refined estimation of channel weights, BAv2 demonstrates significant improvements over the original BAv1 model proposed in our earlier conference publication. The incorporation of an adaptive selection operator in BAv2 enables more effective utilization of features and has been re-evaluated using contemporary training techniques, resulting in substantial performance enhancements across various deep neural network architectures.

This work not only illustrates the versatility of BAv2 in conventional convolutional networks (ConvNets) but also extends its application to the burgeoning field of vision transformers, yielding significant gains in tasks such as image classification, object detection, and instance segmentation. The implications of our findings extend beyond individual architectures, suggesting a broader impact on the field of attention mechanisms within computer vision.

Looking ahead, future research could focus on creating hybrid models that blend the strengths of ConvNets and vision transformers. Additionally, studying how BAv2 integrates with other attention frameworks could help optimize performance. Another important area is to evaluate how well BAv2 works across various applications and datasets, enhancing our understanding of its adaptability and effectiveness. By advancing channel attention mechanisms, our work aims to support the continued development of intelligent visual systems for real-world use. \textcolor{blue}{\textbf{We would like to release our source code.}}

\bibliographystyle{IEEEtran}
\bibliography{reference}

\vfill

\end{document}